\documentclass[conference,a4paper]{IEEEtran}

\IEEEoverridecommandlockouts
\usepackage{cite}
\usepackage{amsmath,amssymb,amsfonts}
\usepackage{algorithmic}
\usepackage{graphicx}
\usepackage{textcomp}
\usepackage{xcolor}
\usepackage{hyperref}

\def\BibTeX{{\rm B\kern-.05em{\sc i\kern-.025em b}\kern-.08em
    T\kern-.1667em\lower.7ex\hbox{E}\kern-.125emX}}

\makeatletter
\let\old@ps@headings\ps@headings
\let\old@ps@IEEEtitlepagestyle\ps@IEEEtitlepagestyle
\def\confheader#1{%
\def\ps@headings{%
\old@ps@headings%
\def\@oddhead{\strut\hfill#1\hfill\strut}%
\def\@evenhead{\strut\hfill#1\hfill\strut}%
}%
\def\ps@IEEEtitlepagestyle{%
\old@ps@IEEEtitlepagestyle%
\def\@oddhead{\strut\hfill#1\hfill\strut}%
\def\@evenhead{\strut\hfill#1\hfill\strut}%
}%
\ps@headings%
}
\makeatother

\IEEEoverridecommandlockouts
\IEEEpubid{\makebox[\columnwidth]
{\hfill 978-1-7281-6218-8/22/\$31.00~\copyright~2022 IEEE}
\hspace{\columnsep}\makebox[\columnwidth]{ }}

\begin{document}

\title{An IoT Based Approach for Platform Independent Positioning Service}

\author{\IEEEauthorblockN{Anh-Van Vu}
\IEEEauthorblockA{\textit{Department of Computer Science} \\
\textit{Korea Advanced Institute of Science and Technology}\\
Daejeon, Republic of Korea \\
vuanhvan@kaist.ac.kr}
\and
\IEEEauthorblockN{Dongsoo Han}
\IEEEauthorblockA{\textit{Department of Computer Science} \\
\textit{Korea Advanced Institute of Science and Technology}\\
Daejeon, Republic of Korea \\
dshan@kaist.ac.kr}

\thanks{This research was supported by Capacity Enhancement Program for Scientific and Cultural Exhibition Services through the National Research Foundation of Korea(NRF) funded by Ministry of Science and ICT (2018X1A3A106860331)}

}

\maketitle
\begin{abstract}
KAIST Indoor Localization System (KAILOS) was introduced in 2014 and became one of the world-first complete indoor positioning solutions. It is mainly based on the WiFi signal to estimate location and provide positioning service via mobile application that has many restrictions by the Operating System. This study proposes an extended architecture of KAILOS to make use of the IoT device for removing the barriers of the mobile-app-based approach and conveniently providing indoor positioning service. Design of the IoT device that contains hardware and software is also presented. Via this study, we prove that the IoT-based approach is promising and can be a crucial feature to widen applications of indoor localization services.
\end{abstract}

\begin{IEEEkeywords}
Indoor Localization, IoT, Wifi-fingerprintting, KAILOS, Crowdsourcing.
\end{IEEEkeywords}

\section{Introduction}

In recent years, the explosive development of the fourth industrial revolution has created enormous demands for inventing new technologies and improving old technologies for industrial and real-world applications. Indoor positioning is one of such technologies. It will provide unprecedented services in many aspects as fueled by the success of outdoor positioning technologies. Indoor positioning technology is expected to be the next significant leap for completing the global positioning system. Therefore, many researchers around the world are focusing on this direction. 

In 2014, KAIST introduced the KAIST Indoor Localization System (\textcolor{red}{\url{https://kailos.io/}}) \cite{han2014kailos}. KAILOS is a complete indoor positioning system consisting of techniques and tools developed for three main components: KAI-Map, KAI-Pos, and KAI-Navi. With KAILOS, it aims at two primary objectives. Firstly, the system can provide indoor localization services to any area where Wi-Fi fingerprinting is available. Secondly, KAILOS provides the service with the required accuracy while maintaining simplicity in building indoor maps and radio maps. With built-in tools, KAILOS allows anyone to contribute to data collection through a standard 6-step process, including Building Registration, Floor Registration, Path and POI Registration, Fingerprint Collection Planning, Fingerprint Collection, and Radio Map Construction. It is known that KAILOS is one of the most accurate WiFi-based indoor positioning systems. As a result, KAILOS has been used in many places since 2014, such as universities, museums, and office buildings. 

Although KAILOS is a complete and stable system, its positioning services are provided to end-users through mobile applications running on Android and IOS, the two most popular operating systems for smartphones globally. On the platform, the user's device must acquire wifi-fingerprinting, bluetooth fingerprinting, and other sensor data for location estimation. However, the access to the signals is limited partially or totally by the OS. For example, in the IOS operating system, the application is not allowed to scan WiFi signals from the surrounding environment, while the Android OS provides an API for this purpose with restrictions on scanning frequency and scanning time (both foreground and background scan) that may cause a serious problem with consistent localization. Further, the installation of the application causes inconvenience for occasional users and one-time users. These problems significantly limit the scalability of KAILOS in fields.AP

To eliminate the dependence on operating systems and enrich the applications of KAILOS, this paper presents a new approach based on Internet of Things (IoT). Accordingly, we propose an extended architecture of KAILOS system with the participation of specially designed wearable devices that assist smartphone users in accessing KAILOS services regardless of the operating system. The IoT device is equipped with multiple connectivities and sensors, including LTE, WiFi, BLE,  GNSS, light sensor, pressure sensor, and inertial sensor. By being interfered deeply to the physical layer, the device allows continuous data collection at high sampling rates to feed the KAILOS engine for estimating the precise location. Users can easily access locating services via mobile web browsers by scanning a barcode printed on the IoT device without installing any smartphone application. The approach based on IoT enables a plug-and-play service.

In the remaining part of the paper, we will briefly describe the related works (Section II), then focus on explaining the overall structure of the extended architecture (Section III), and the design of the IoT device (Section IV). In Section V, we will present the experimental setup and results of the evaluation. The final part of the paper is Section VI, where we draw our conclusion with the limitation of the study and future works.

\section{Related Works}
In the history of indoor positioning, numerous studies have been conducted. Most of these studies are based on well-known wireless technology such as WiFi, Ultra Wide Band (UWB), Bluetooth, Ultrasound, and Zigbee \cite{zafari2019survey}. 

The latest version of Bluetooth technology is Bluetooth Low Energy (BLE). Thanks to the advantages of throughput, range, and low energy consumption, BLE is widely used in many personal devices. Studies on the application of BLE positioning aim to take advantage of this popularity. Theoretically, upon the properties of BLE, the applicable techniques can be RSSI, AoA, and ToF \cite{jianyong2014rssi, monfared2018experimental, giovanelli2018bluetooth}. Nevertheless, existing BLE-based localization systems primarily utilize RSSI. In addition, the operation of BLE in 2.4GHz and 5Ghz frequency under low power mode makes it pretty sensitive to interference. These reasons seriously limit the accuracy of the BLE-based localization. Consequently, it is only used in proximity and context detection applications. iBeacon (by Apple) and Eddystone (by Google) are two typical BLE-based protocols designed for the above applications.

Ultra Wide Band (UWB) technology is particularly optimized to the physical layer for short-range localization based on the received signal strength level. UWB employs an ultra-wide frequency band from 3.1GHz to 10.6GHz to transmit ultra-short pulses (period of less than one ns). This technology is beneficial for minimizing power consumption, ensuring resistance to interference, and increasing the penetration of materials such as cement, steel, and even liquids \cite{zhang2006accurate, stelios2008indoor, poulose2020uwb}. Although this technology is ubiquitous in high-end handheld devices, the UWB-based positioning system requires installing assistive devices in the target spaces. This installation is not only expensive but also affects the aesthetics and the layout of the building. Hence, these disadvantages are major barriers to building a general indoor positioning system on a large scale.

Like UWB, Ultrasound-based localization can also provide relatively high accuracy in certain conditions. This positioning technology is based on the understanding of sound waves propagation in a physical environment \cite{weaver1990anderson, hoeflinger2019passive, kuban2005ultrasound}. We can indirectly calculate the distance between objects through physical formulas by determining sound waves' velocity in a specific environment and measuring propagating time using ultrasound receivers and transmitters. However, the most significant disadvantage of this method originates from the characteristics of propagating environment, which frequently change with the variance of temperature and humidity. Dealing with this problem requires additional subsystems to measure environmental conditions and compensate for the errors. Besides, ultrasound receivers and transmitters are rather complex and cumbersome, which practically limits the applications of this technique.

In addition to the above technologies, Radio Frequency Identification Device (RFID) \cite{hahnel2004mapping, xiao2017one, bouet2008rfid}, Zigbee \cite{bianchi2018rssi, sugano2006indoor, blumenthal2007weighted}, Visible Light (VL) \cite{luo2017indoor, zhou2012indoor, zhang2014asynchronous}, and Acoustic Signal \cite{sertatil2012novel, huang2014swadloon} have also been studied for indoor positioning. However, they are less common for several reasons, and two primary reasons are high cost and low scalability.

WiFi-based indoor localization has been studied quite early \cite{han2014building, zafari2019survey, kotaru2015spotfi, lim2007real, lee2012voronoi, lim2013radio}. Compared with other techniques, it can provide location service at a low cost due to the abundantly global availability of APs and mobile devices. For this reason, WiFi-based technology is expected to become the core of a global indoor positioning system. Indeed, in recent years, the world's foremost technology companies such as Google, Apple, and Microsoft have begun to exploit the abundance of WiFi signals on the global scale to provide indoor location services combined with GPS, Glonass, Beidou, Galileo, etc. Although the current accuracy and performance are not impressive due to the passive data collection method, the participation of these big-tech companies in studying WiFi-based locating technologies will definitely create an enormous opportunity to promote the development of this field.

\section{Overall Structure of the System}

\begin{figure*}[t]
  \centering
  \includegraphics[width=125mm]{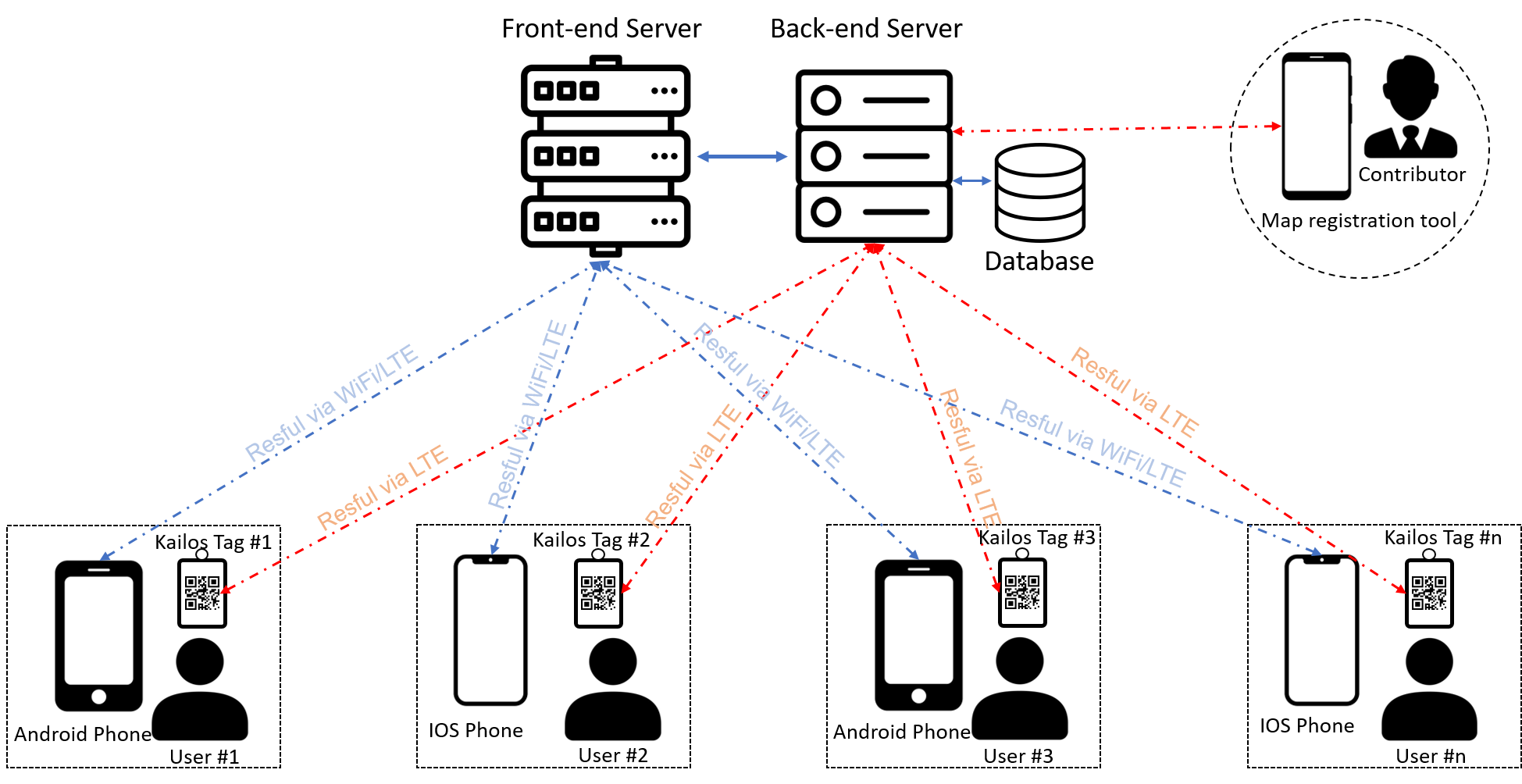}
  \caption{Overall Structure of the IoT based Approach}
  \label{fig:overall_Structure}
\end{figure*}

Unlike the conventional positioning system, where the mobile application directly collects data, interacts with the back-end server, and visualizes location, the extended system involves the IoT devices. Despite the change in the overall architecture, the extended architecture entirely inherits the communication interface between the end devices and the server, thus ensuring backward compatibility with the traditional mobile applications. The overall architecture of the system is shown in Fig \ref{fig:overall_Structure}, which includes the main components of the system, such as IoT device - smartphone pair, Front-end Server, Back-end Server, Database, and Map Registration tools. Details of each component will be presented as below.

\subsection{Front-end and Back-end Server}\label{AA}

\begin{figure}[b]
  \centering
  \includegraphics[width=85mm]{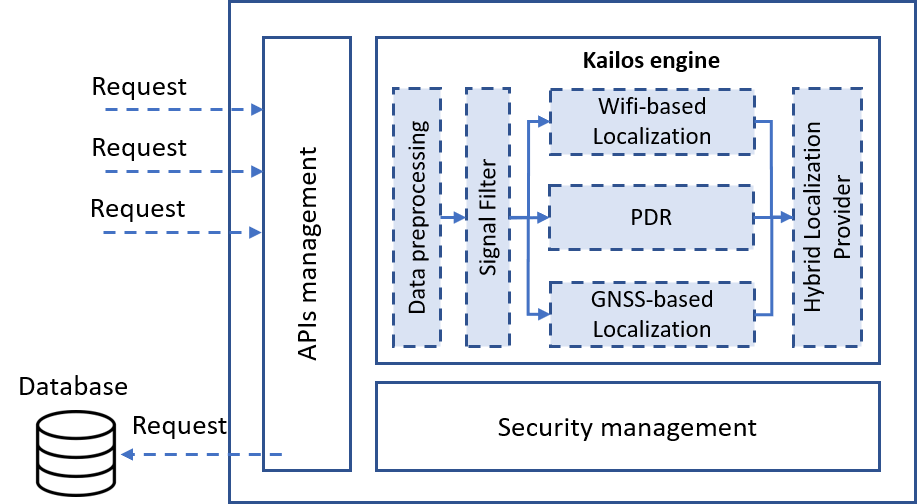}
  \caption{Back-end Server Diagram}
  \label{fig:back-end_diagram}
\end{figure}

The back-end server is the most crucial element of the KAILOS system, which manages the operation of other components in the system. Thanks to well-designed APIs, the back-end server can flexibly interact with subsystems and other systems without any difficulty. Moreover, the back-end server also incorporates KAILOS engine, which is the heart of the indoor positioning system. KAILOS engine includes a set of filters, wifi-based location estimator, sensor fusion, pedestrian dead reckoning, hybrid localization, and other advanced data processing algorithms (Fig.\ref{fig:back-end_diagram}). On the other side, the Front-end server is mainly responsible for user interface and position visualization. Whenever a request from a user's smartphone is made, the front-end server will immediately connect to the back-end side to inquire recorded location of the associated IoT device and manage the visualization via mobile web browsers. In this way, multiple smartphones can subscribe to the location information of a single IoT device, which efficiently benefits group-user in many applications.

\subsection{Database}

If KAILOS engine is the heart of the system, the database can be considered as the blood, which is also an indispensable part. The core of this database stores a large number of indoor maps, radio maps, and sensor data collected from end devices. The remaining part of the database is for supporting users and crowdsourcing contributors to build the map. Within the database, data is classified and stored systematically by area, building, floor, and floor paths. This storage is beneficial for visualization and provides essential information for smooth transitions between indoor and outdoor environments.

\subsection{IoT Device and Smartphone Pair}

\begin{figure}[!hb]
  \centering
  \includegraphics[width=80mm]{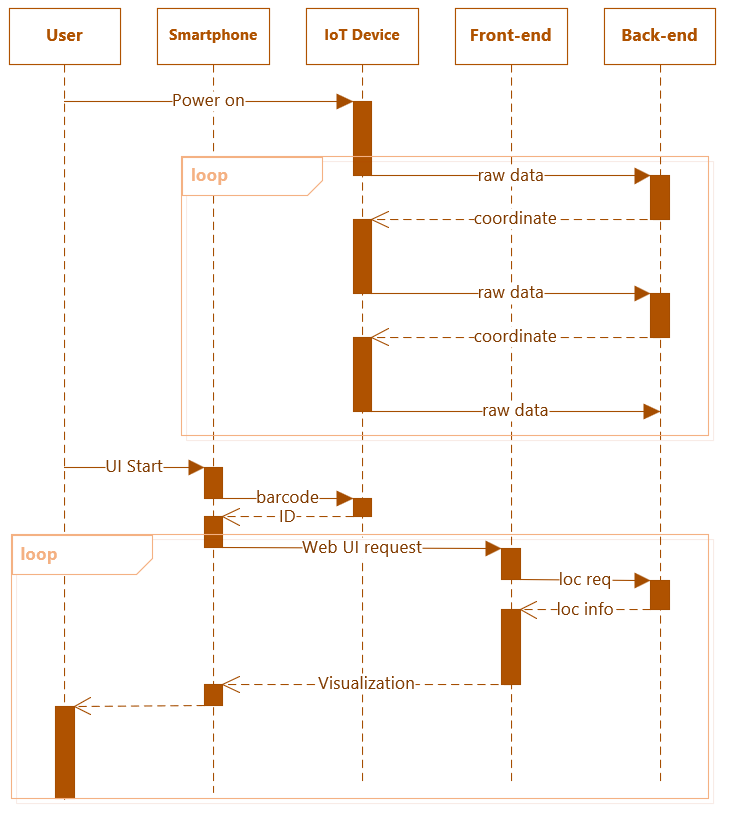}
  \caption{Interaction of the IoT Device with other objects of the system}
  \label{fig:Sequence_Diagram}
\end{figure}

IoT device and smartphone are paired for the users to access positioning services. In the conventional approach, we operated an Android application to simultaneously visualize locations and collect input data for the location estimator. The limitation of this manner comes from the operating system because it prohibits applications from getting WiFi signals at a high scanning rate. Starting from Android 9, the foreground apps can scan only four times every 2 minutes, and all belonging background apps can scan once per 30 minutes \cite{google_restriction}. As a result, the control of sampling frequency significantly affects the performance of the positioning service. One of the workarounds to this problem is to use interpolation techniques. However, this solution is unsuitable for applications requiring continuous positioning with high accuracy. On the other hand, with the participation of an IoT device, this issue can be entirely handled by dividing the tasks between the smartphone and IoT device. The smartphone is used only for visualization and user interface, while the IoT device is responsible for data collection. This strategy removes the restrictions of the operating system and provides high-quality data for the location server. It is especially helpful because the iOS operating system completely prohibits scan of WiFi signals from the applications.

Fig.\ref{fig:Sequence_Diagram} illustrates the operation of the smartphone and IoT device in association with other objects of the system. For each IoT device,  we assign an unchangeable UUID. When the power is on, it continuously collects the necessary data, including wifi-fingerprinting, IMU data, and GNSS data, and sends them to the backend server. KAILOS engine then uses these data to estimate the position of the device and store this information on the backend server. When the users request the positioning service, smartphones should be used to scan the barcode printed on the device to acquire an URL that contains the UUID of the IoT device. This URL will automatically navigate the user to a web interface that displays the location and other relevant service information. In the next section of the paper, we will introduce the design of the IoT device and its details, including hardware and software.

\section{Design of the IoT Device}

\subsection{Hardware Design}

\begin{figure}[b]
  \centering
  \includegraphics[width=85mm]{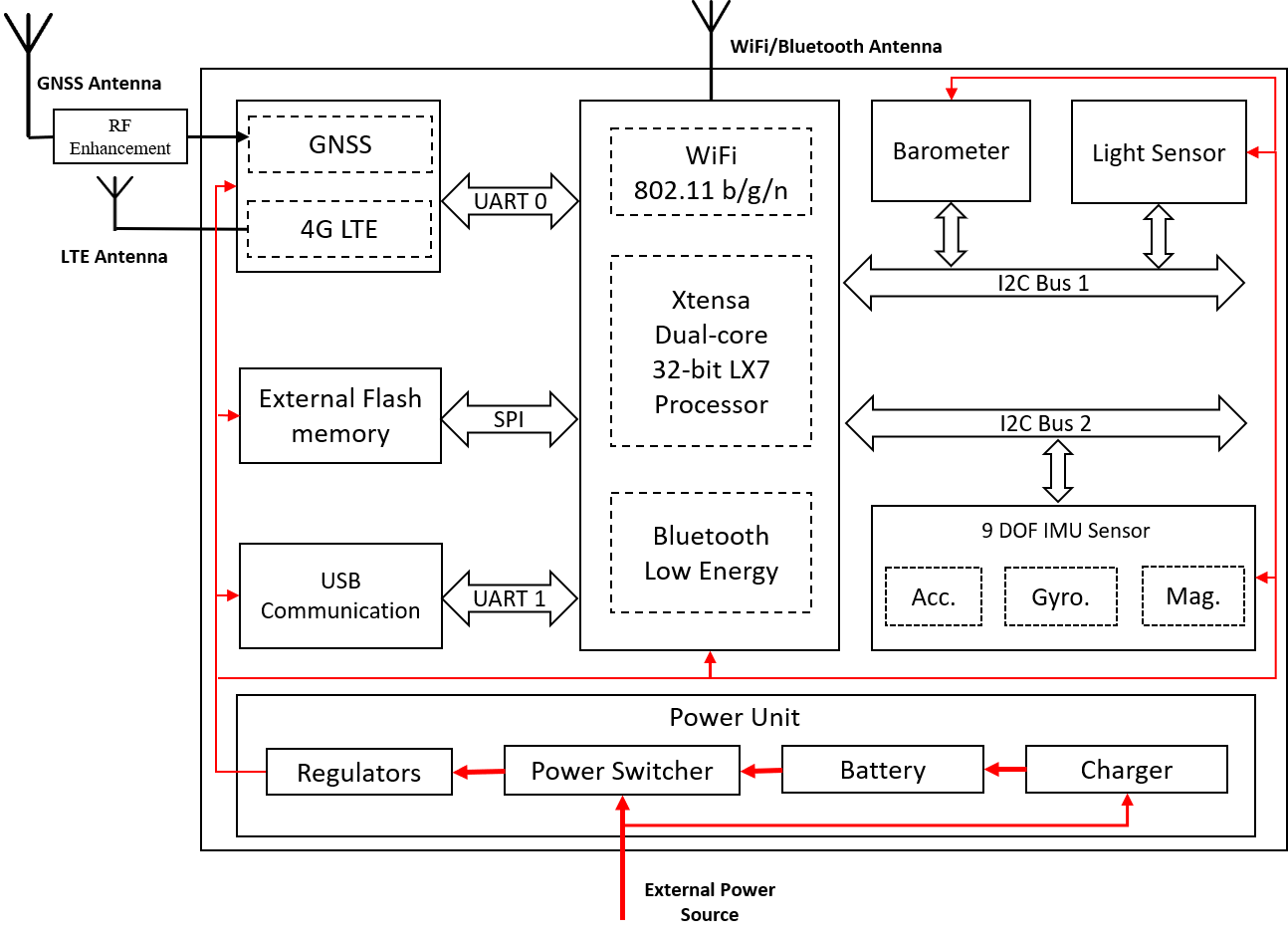}
  \caption{Block diagram of IoT device design}
  \label{fig:IoT_Device_Diagram}
\end{figure}

We design an IoT device with processing Unit (MCU), WiFi, 4G LTE, BLE, IMU sensor, GNSS, Flash memory, and USB communication. In addition, there are different sensors, such as a light intensity sensor for Indoor/Outdoor context switching detection and a barometer for floor detection. The block diagram of the IoT device is shown in Fig.\ref{fig:IoT_Device_Diagram}. The central processor of the device is a 32-bit MCU that is in charge of coordinating the operation of the remaining components. The advance of System On Chip (SoC) technology allows us to select hardware integrating MCU, WiFi, and BLE on a single chip. In this way, design complexity is reduced while energy efficiency is significantly increased. The same strategy is applied to choose one SoC, which encloses both GNSS and LTE, and another SoC for Accelerometer, Gyroscope, and magnetometer. To connect the components of the system with the MCU, we mainly use three communication standards, including universal asynchronous receiver-transmitter (UART), Serial Peripheral Interface (SPI), and Inter-Integrated Circuit (IIC). Among the above communication standards, UART is used to connect to LTE/GNSS SoC, SPI is used to connect to External Flash Memory, and I2C is used to connect to other sensors. Theoretically, each I2C bus allows connecting to multiple peripherals and sensors. However, we use two distinct I2C buses for two groups of sensors classified by the sampling rate in this design. Usually, the barometer has a lower sampling rate than IMU and other sensors. Consequently, if they are all connected on a single bus, the sampling rate of the IMU sensor will be capped by the low sampling rate sensors. This phenomenon remarkably affects the quality of the positioning and heading estimation algorithm.

\begin{figure}[!b]
  \centering
  \includegraphics[width=75mm]{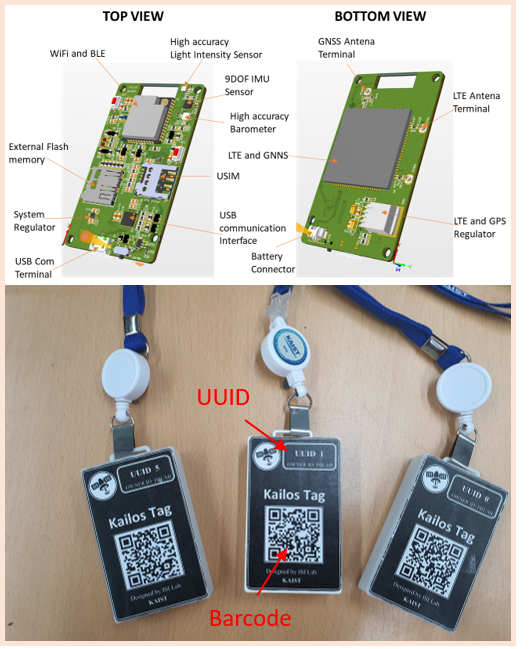}
  \caption{Hardware design and appearance of the real IoT Device}
  \label{fig:real_IoT_Device}
\end{figure}

\setlength{\tabcolsep}{9pt}
\renewcommand{\arraystretch}{1.3}
\begin{table}[!t]
\caption{Specification of the IoT device}
\label{tab:device_spec}
\begin{tabular}{|l|l|}
\hline
\textbf{CPU}          & Xtensa  32-bit LX6 240 MHz                                                                                                                          \\ \hline
\textbf{WiFi}         & 802.11 b/g/n                                                                                                                                        \\ \hline
\textbf{Bluetooth}    & V4.2  BR/EDR and BLE                                                                                                                                \\ \hline
\textbf{LTE}          & LTE-TDD/ LTE-FDD/HSPA+/GSM/GPRS/EDGE                                                                                                                \\ \hline
\textbf{GNSS}         & GPS, GLONASS, BEIDOU, GALILEO, QZSS                                                                                                                 \\ \hline
\textbf{IMU Sensor}   & \begin{tabular}[c]{@{}l@{}}Accelerometer: ±2g, ±4g, ±8gand ±16g\\ Gyroscope: ±250, ±500, ±1000, and ±2000°/sec\\ Magnetometer: ±4800 \(\mu T\)\end{tabular} \\ \hline
\textbf{Light Sensor} & 0 - 120000 lx                                                                                                                                       \\ \hline
\textbf{Barometer}    & 10 - 2000 mbar (resolution 0.016 mbar)                                                                                                              \\ \hline
\textbf{Flash Memory} & Up to 128GB                                                                                                                                         \\ \hline
\textbf{Battery}      & Li-Ion 900 mAh                                                                                                                                      \\ \hline
\end{tabular}
\end{table}

In any wearable device, the power unit is also an essential part, which supplies power to the whole system. In this design, the system power is supplied by a rechargeable battery via a linear voltage regulator. Although linear voltage regulators are slightly less efficient than switching regulators, they have outstanding stability. The linear regulators do not interfere with surrounding circuits by the high-order harmonics that are regularly produced by the high-frequency switching operation in the switching regulators. In addition, a power switcher and a charger are also included to automatically switch between power sources and charge the battery when a connection to an external power source is detected.

The hardware design of the IoT devices and its specification is described in Fig.\ref{fig:real_IoT_Device} and Table \ref{tab:device_spec}. The IoT devices are assembled in the form of wearable tags, and each device is assigned an unique UUID, Barcode, and NFC tag. The users can either utilize the smartphone camera for scanning the barcode or tapping the smartphone NFC reader to quickly access the visualization of positioning services. A device could work properly with an additional sim card. Each sim card should be registered to the local cellular network operators in advance to be able to connect to the internet via LTE or GPRS. The devices can automatically switch between LTE and GPRS mode to maintain the connection. However, the LTE mode has a higher priority for energy-saving purposes.

\subsection{Software Design}

\begin{figure}[b]
  \centering
  \includegraphics[width=80mm]{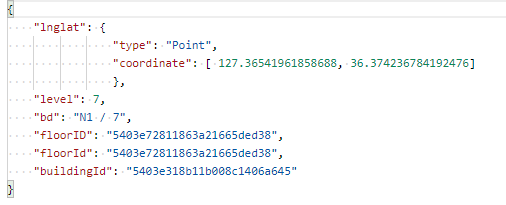}
  \caption{Response from server}
  \label{fig:response}
\end{figure}

\begin{figure}[!t]
  \centering
  \includegraphics[width=70mm]{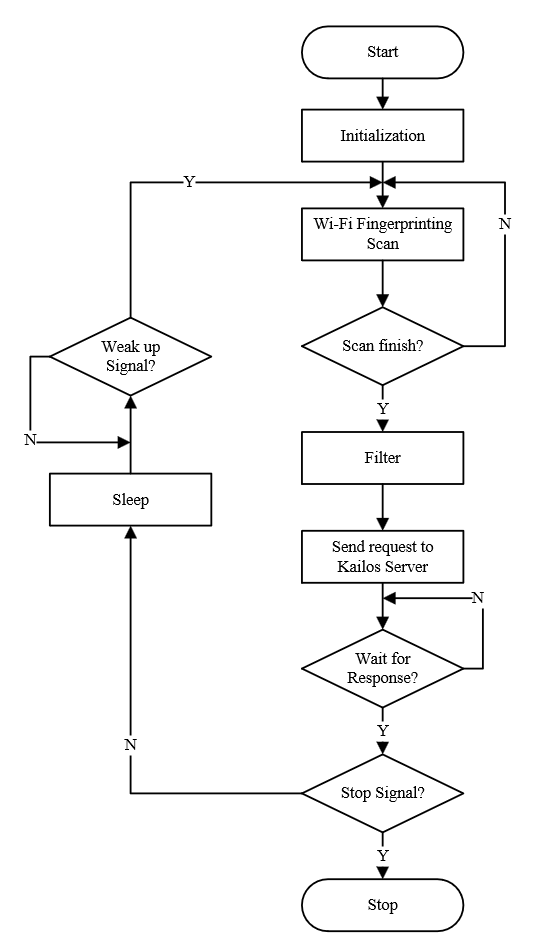}
  \caption{Algorithms of the IoT device}
  \label{fig:iot_device_algorithms}
\end{figure}

KAILOS is a hybrid positioning system that combines WiFi, Bluetooth, and PDR positioning techniques. In the hybrid positioning system, each positioning technique is implemented independently and then integrated together. Therefore the system can work without the availability of the bluetooth signal,  and/or IMUs signal. In this study, we aim to make the IoT devices work independently with the WiFi-based positioning technique that is the core technology of KAILOS system. The flow chart in Fig.\ref{fig:iot_device_algorithms} explains the algorithms of the IoT device. There are three phases in the operation of the IoT device, including Initializing Phase, Positioning Phase, and Sleeping Phase. In the initializing phase, the device executes configuration to ensure proper operation of the LTE, WiFi, Bluetooth, and sensors. Once the initialization is done, the device immediately enters the positioning phase that scans the available WiFi signals from the surrounding environment and sends it to the server via LTE to estimate the location. This phase is finished either if the device receives a response from the server or if a timeout happens. The sleeping phase is designed to prolong the battery life of the device. Putting the device in sleep mode will minimize the power consumption, leading to a longer-time operation of the device. This sleeping phase starts at the end of the positioning phase and lasts until a low power timer triggers a wake-up signal.

\begin{figure}[t]
  \centering
  \includegraphics[width=70mm]{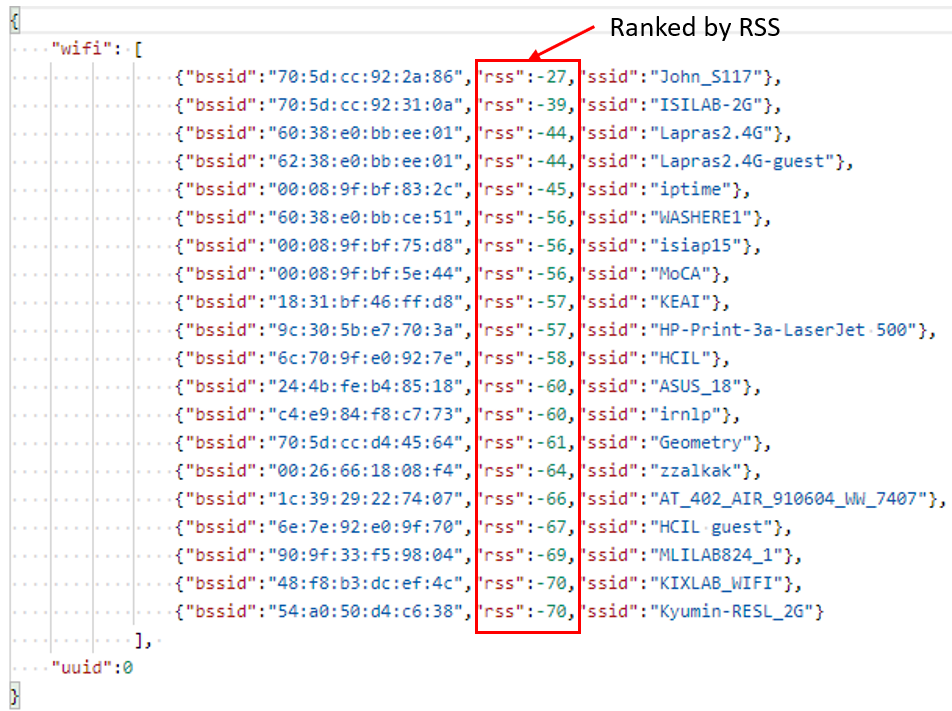}
  \caption{Data body of the request}
  \label{fig:data_body}
\end{figure}

In the positioning phase, instead of directly sending the raw data to the server, we apply some pre-processing procedures to normalize it. After the scanning process, the device obtains a list of available APs, and each AP has a set of information. We detect and remove the abnormal APs from the list by checking the missing information of each AP. For the remaining APs, we rank the APs by the Receive Signal Strength (RSS), then select the top 20 APs accompanied with the device UUID to generate the data body of the request. If the number of detected APs is less than 20, all of them are included. By this procedure, we can enhance the robustness of wifi-fingerprinting and estimation results. Moreover, removing unnecessary data is significantly beneficial for saving the cost of data plan. The Fig.\ref{fig:data_body} presents the data body of a request. After obtaining the wifi fingerprinting from the IoT device, KAILOS server takes several milliseconds to infer and return the device's location. The location data contains longitude, latitude, building ID, floor ID, building name, and floor name as specified in Fig.\ref{fig:response}.

\section{Experiments and Results}

\subsection{WiFi Based Indoor Localization}

To evaluate the performance of the new approach, we conducted an experiment on the campus of Korea Advanced Institute of Science and Technology (KAIST), Daejeon, Republic of Korea. Within the KAIST campus, radio maps of most buildings are available in KAILOS database. Thus, we could set up the experiment without so much difficulty. The IT building(N1), where our office is located was selected as the target building. N1 is a building located on the north side of the campus. Table \ref{tab:it_building_information} depicts some information about the building.

\setlength{\tabcolsep}{9pt}
\renewcommand{\arraystretch}{1.2}
\begin{table}[!b]
\caption{The information of the IT building N1}
\label{tab:it_building_information}
\begin{tabular}{|l|c|}
\hline
\textbf{Building Name}                                                                                      & IT building N1                \\ \hline
\textbf{Building ID}                                                                                        & 5403e318b11b008c1406a645      \\ \hline
\textbf{\begin{tabular}[c]{@{}l@{}}Lower Left Corner  Coordinate \\ {[}longitude, latitude{]}\end{tabular}} & {[}127.3652157, 36.3740271{]} \\ \hline
\textbf{\begin{tabular}[c]{@{}l@{}}Upper Right Corner Coordinate\\ {[}longitude, latitude{]}\end{tabular}}  & {[}127.3662484, 36.3743834{]} \\ \hline
\textbf{Length (m)}                                                                                         & 114.6                         \\ \hline
\textbf{Width (m)}                                                                                          & 39.6                          \\ \hline
\textbf{Number of Floor}                                                                                    & 9                             \\ \hline
\textbf{Number of Basement}                                                                                 & 1                             \\ \hline
\end{tabular}
\end{table}

\begin{figure}[t]
  \centering
  \includegraphics[width=64mm]{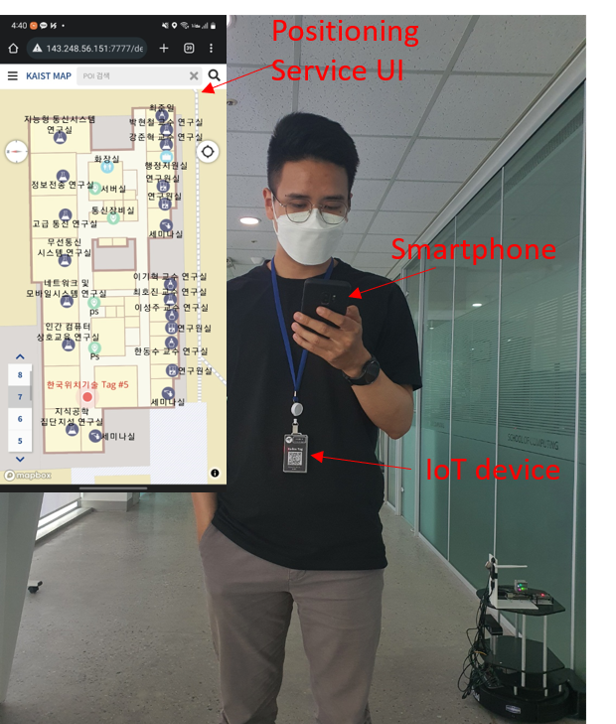}
  \caption{Experimental setup for the valuation}
  \label{fig:experimental_setup}
\end{figure}

\begin{figure}[t]
  \centering
  \includegraphics[width=90mm]{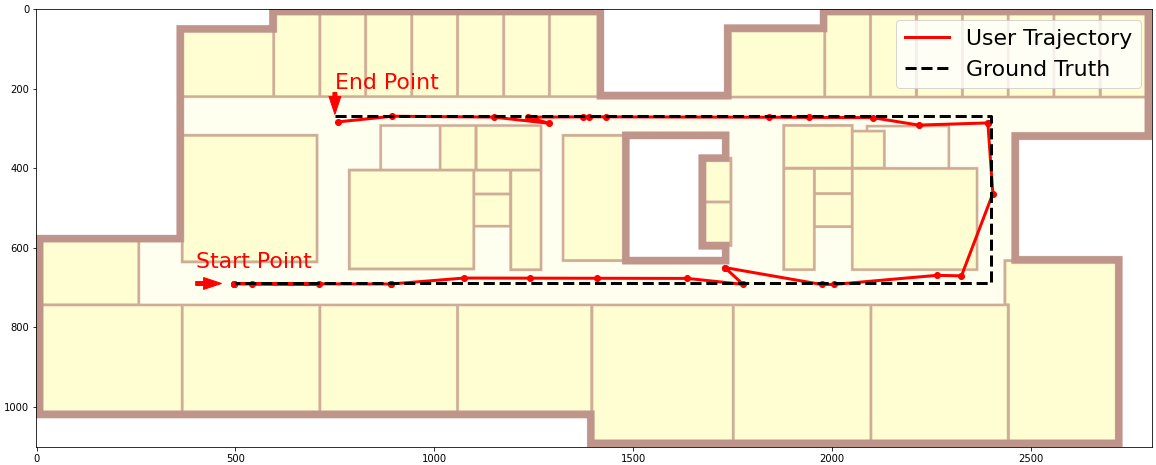}
  \caption{The trajectory of the device compared with ground truth}
  \label{fig:result1}
\end{figure}

Fig.\ref{fig:experimental_setup} shows the setup of the experiment where the accuracy was measured while a person wearing the IoT device was moving along the hallway of each level. The real-time visualization of the location is done by using a Samsung Galaxy Note 10+ and an iPhone 11 pro. Fig.\ref{fig:result1} compares the estimated trajectory of the device and the ground-truth trajectory on the 7th floor of the building. As shown in the figure, the estimated trajectory is very close to the ground-truth one. 

\setlength{\tabcolsep}{4pt}
\renewcommand{\arraystretch}{1.2}
\begin{table}[!t]
\caption{Comparison of IoT based Approach and mobile-app based Approach}
\label{tab:compare_smartphone}
\begin{tabular}{|l|c|c|c|}
\hline
                                                                          & \textbf{\begin{tabular}[c]{@{}c@{}}IoT Device \\ \& Smartphone\end{tabular}} & \multicolumn{1}{l|}{\textbf{Note 10+ Plus}} & \multicolumn{1}{l|}{\textbf{iPhone 11 Pro}} \\ \hline
\textbf{Smartphone OS}                                                    & Android or IOS                                                               & Android                                     & IOS                                         \\ \hline
\textbf{\begin{tabular}[c]{@{}l@{}}Free \\ App-Installation\end{tabular}} & Yes                                                                          & No                                          & not available                               \\ \hline
\textbf{Group User}                                                       & Yes                                                                          & No                                          & not available                               \\ \hline
\textbf{Time per scan (s)}                                                & 2                                                                            & 4                                          & not available                               \\ \hline
\textbf{\#Scan per minute}                                                 & 30                                                                           & 2                                         & not available                               \\ \hline
\textbf{Average Error (m)}                                                & 2.4                                                                          & 2.5                                         & not available                               \\ \hline
\end{tabular}
\end{table}

To compare the IoT base approach and the mobile app-based approach, we also measured the performance of the smartphone app. Typically, the iOS operating system disallows scanning WiFi from applications leading to the unavailability of the iPhone app. Therefore, the comparison was made mainly between the IoT device and the android smartphone app. Following Table \ref{tab:compare_smartphone}, we can conclude that the performance of the IoT device is better than the android app in terms of accuracy, scanning frequency, and time per scan. Especially, non-OS-constraint, Free App-Installation, and Group User are the advanced features of the IoT-based approach compared with the conventional one.

\subsection{Effect of Sampling Rate on Pedestrian Dead Reckoning}

\begin{figure}[b]
  \centering
  \includegraphics[width=90mm]{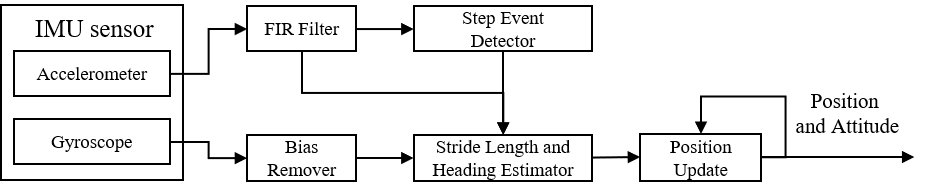}
  \caption{Pedestrian Dead Reckoning Algorithms}
  \label{fig:pdr}
\end{figure}

Besides the benefit of using IoT based approach to WiFi-based positioning, collecting sensing data at high sampling frequencies can enhance the performance of sensor-fusion positioning techniques. In this experiment, we collected the IMU sensor data at different sampling rates, then applied a PDR algorithm and compare the results produced by different input data.

The pipeline of the used PDR algorithms in this experiment is shown in Fig.\ref{fig:pdr}. The algorithm uses Accelerometer and Gyroscope data to estimate the stride length and orientation of the foots to present the object's positions. This algorithm has the benefit of being simple and not requiring foot-mounted sensors to produce a good result. 
For step event detection, we mainly used acceleration data with some basic signal processing techniques. Firstly, we calculated the magnitude of the acceleration before applying a low pass filter and removing the effect of gravity. Secondly, we binarized the signal and recognized the stance phase and swing phase of the steps by monitoring the raising and lowering pulses.

Theoretically, the heading information can be obtained in some different ways. For example, we can calculate absolute orientation from acceleration and magnetic field or retrieve relative orientation from the gyroscope. The optimal weighting algorithms such as Mahony or Madwick can also be an option \cite{madgwick2010efficient, mahony2008nonlinear}. In this case, we use angular velocity signal from the gyroscope and acceleration signal from the accelerator to represent the device's orientation via direction cosine matrix (DCM) \cite{woodman2007introduction, jimenez2009comparison}:

\begin{equation}
    C(t+\delta t) = C(t).exp\left (\int_{t}^{t+\delta t}\omega(\tau)d\tau)\right )
    \label{eq:dcm}
\end{equation}

where \(C(t+\delta t)\) and \(C(t)\) are the rotation matrix at time step \(t+\delta t\) and \(t\) respectively, and \(\omega(\tau)\) is instantaneous angular velocity.

The stride length should also be estimated at every detected step. To this end, we imitate a method proposed by Weinberg of Analog Device \cite{weinberg2002using}. This method aims to exploit the relationship between stride length and the bounce, which actually is the vertical movement of the human hip while walking. By doing so, the technique can robustly produce the distance measurement at an accuracy of 92 \% over a variety of subjects of different leg lengths. 

\begin{equation}
    S = \sqrt[4]{A_{max} - A_{min}}.n.K 
    \label{eq:stride_length}
\end{equation}

Equation \ref{eq:stride_length} shows how the distance is estimated via acceleration information. In that, \(S\) is walking distance (equal to stride length if the number of steps walked \(n=1\)), \(A_{min}\) and \(A_{max}\)  are  the  minimum  and maximum acceleration  measured  in  the  Z axis in a single stride respectively. \(K\) is a constant for unit conversion (i.e., feet or meters traveled)

Consequently, the coordinate \(P(x_k, y_k)\) of the object at detected step \(k\) can be determined as:

\begin{equation}
    \left\{\begin{matrix}
    y_k = y_{k-1} + S_k.\cos(\delta_k)\\
    x_k = x_{k-1} + S_k.\sin(\delta_k)
    \end{matrix}\right.
\end{equation}

where \(S_k\) and \(\delta_k\) are stride length and orientation at stance phase of step \(k\) respectively.

\begin{figure}[t]
  \centering
  \includegraphics[width=78mm]{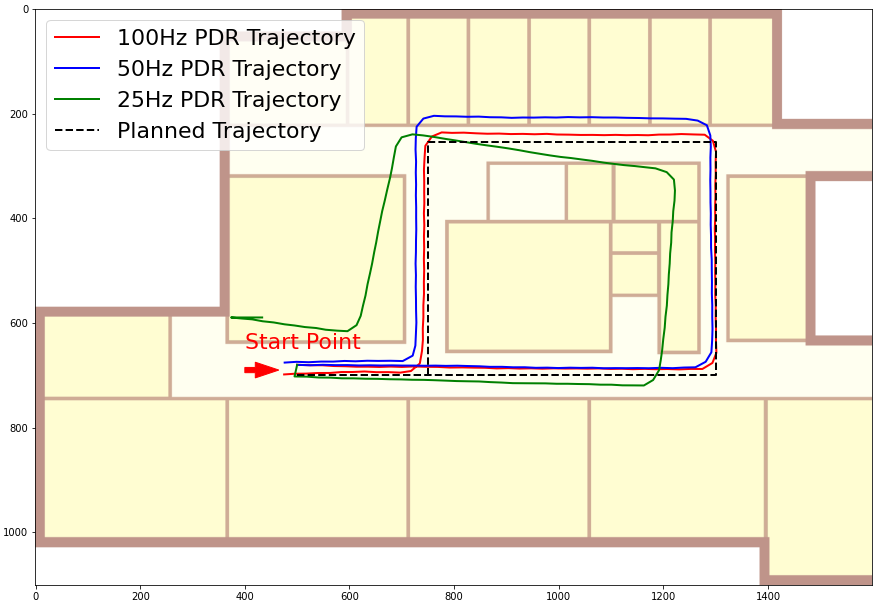}
  \caption{Performance of the PDR Algorithm in different sampling rates}
  \label{fig:compare_pdr}
\end{figure}

\setlength{\tabcolsep}{5pt}
\renewcommand{\arraystretch}{1.5}
\begin{table}[t]
\caption{The sampling rate of the sensors of different platforms}
\label{tab:compare_freq}
\begin{tabular}{|l|c|c|c|c|}
\hline
                       & \textbf{IoT Device} & \multicolumn{1}{l|}{\textbf{Note 10+ Plus}} & \multicolumn{1}{l|}{\textbf{Galaxy S6}} & \textbf{iOS} \\ \hline
\textbf{Accelerometer} & 1000 Hz             & 500 Hz                                      & 200 Hz                                  & 100 Hz       \\ \hline
\textbf{Gyroscope}     & 1000 Hz             & 500 Hz                                      & 200 Hz                                  & 100 Hz       \\ \hline
\textbf{Magnetometer}  & 100 Hz              & 100 Hz                                      & 100 Hz                                  & 100 Hz       \\ \hline
\end{tabular}
\end{table}

Similar to the previous experiment, we selected the N1 building for data collection. This time, we collected the IMU sensor data at three different sampling rates (25Hz, 50Hz, and 100Hz) along the same planned trajectory on the 7th floor of the building. Fig. \ref{fig:compare_pdr} shows the trajectories estimated by the PDR algorithm for 25Hz, 50Hz, and 100Hz data. Compared with the ground-truth trajectory, the trajectory corresponding to 100 Hz data has the best matching, followed by the trajectory of 50Hz data, while 25 Hz data result in the least matching trajectory. The experimental result demonstrated that the higher sampling rate resulted in a better performance of the positioning algorithm. Table \ref{tab:compare_freq} also indicates that the IoT device outperforms other mobile platforms in terms of maximum sampling rate of the integrated sensors.
Further, the IoT device approach provides not only high-speed data collection but also a mostly perfect sampling period that significantly benefits the data-preprocessing step.

\section{Conclusion and Discussion}

This paper presented an IoT-based approach to a WiFi-based indoor positioning system. To our knowledge, this is the first time an IoT device is included in a system such as KAILOS to eliminate the weaknesses caused by providing service on mobile platforms. The results demonstrated that the approach is promising and will be a great opportunity to popularize applications of indoor positioning in real life. Our current deployment of the KAILOS system with the IoT devices in a Science Museum to analyze the visitor's behaviors is one of examples.

In a nutshell, there are many advantages when a dedicated positioning hardware device equipped with various sensors is prepared. First, it can free us from the limitations of mobile platforms. WiFi signals and GPS signals can be collected whenever necessary. Second, the signal collection period of the sensors can be arbitrarily adjusted, and each sensor can have a different signal collection cycle. By collecting signals according to the signal collection cycle of each sensor, it is possible to prevent some signals from being dropped. In addition, it is possible to develop an optimal sensor access protocol for various sensors. Third, it is possible to develop a sensor fusion technique by installing new sensors that are not available in smartphones. For example, some environmental sensors (for indoor/outdoor detection). If necessary, these sensors can be easily included in the device. Although the current design has some limitations, such as it doesn't work with 5GHz WiFi, we consider this study is the very first step of our ambitiousness to produce an indoor/outdoor positioning SoC for any wearable device. For future works, we are going to implement the full functionality of the IoT device for the extended KAILOS system. Additionally, we will improve the hardware design to make it more lightweight, have a longer battery life, and support dual-band WiFi (2.4GHz and 5GHz).

\bibliographystyle{IEEEtran}
\bibliography{reference}

\end{document}